\documentclass[10pt,twocolumn,letterpaper]{article}

\usepackage{cvpr}              
\definecolor{cvprblue}{rgb}{0.21,0.49,0.74}
\usepackage[pagebackref,breaklinks,colorlinks,allcolors=cvprblue]{hyperref}

\usepackage[normalem]{ulem}
\useunder{\uline}{\ul}{}
\usepackage{multirow}
\usepackage{tabularx}
\usepackage{pifont}
\usepackage{multirow}
\usepackage{makecell}
\usepackage{threeparttable}

\newcommand{\cmark}{\ding{51}}%
\newcommand{\xmark}{\ding{55}}%
\newcommand{\nx}{\multicolumn{1}{c}{\xmark}}
\newcommand{\na}{\multicolumn{1}{c}{--}}

\def\paperID{29} 
\def\confName{CVPR}
\def\confYear{2026}

\title{PointTransformerX: \\Portable and Efficient 3D Point Cloud Processing without Sparse Algorithms}

\author{Laurenz Reichardt \quad Nikolas Ebert \quad Oliver Wasenm\"uller\\
Mannheim University of Applied Sciences, Germany\\
{\tt\small \{l.reichardt, n.ebert, o.wasenmueller\}@th-mannheim.de}
}

\begin{document}
\maketitle
\begin{abstract}
3D point cloud perception remains tightly coupled to custom CUDA operators for spatial operations, limiting portability and efficiency on non-NVIDIA, AMD, and embedded hardware. We introduce \textbf{PointTransformerX} (PTX), a fully PyTorch-native vision transformer backbone for 3D point clouds, removing all custom CUDA operators and external libraries while retaining competitive accuracy. PTX introduces 3D-GS-RoPE, a rotary positional embedding that encodes 3D spatial relationships directly in self-attention without neighborhood construction, and further replaces sparse convolutional patch embedding with a linear projection. PTX explores inference-time scaling of attention windows to improve accuracy without retraining. With a redesigned feed-forward network, PTX achieves 98.7\% of PointTransformer V3's accuracy on ScanNet with \textbf{79.2\% fewer parameters} and executing \textbf{1.6$\times$ faster} while requiring just \textbf{253 MB memory}. PTX runs natively on NVIDIA GPUs, AMD GPUs (ROCm), and CPUs, providing an efficient and portable foundation for point cloud perception.
\end{abstract}
    
\section{Introduction}

Point cloud perception is fundamental to a wide range of applications, including robotics, industrial automation, autonomous driving, and augmented and virtual reality.
However, point cloud data is spatially unordered, and neural networks therefore rely on explicit spatial algorithms such as neighborhood grouping, voxelization, and radius search to extract geometric features, while other approaches rely on improved data coverage \cite{text3daug}.
Such algorithms are computationally and memory intensive, and in practice neural networks almost exclusively integrate custom CUDA operators and dedicated libraries \cite{STFormer,Minkowski,Torchsparse,spconv,PYG,pointops,SparseConvNet}.
This creates a structural dependency on high-end NVIDIA hardware, preventing deployment on hardware from other vendors.
Furthermore this relies on a fragmented library ecosystem which is often not maintained and lags well behind the standard PyTorch framework.
As a result, modern capabilities such as BF16 and FP8 precision formats, JIT compilation, and kernel fusion are poorly supported across the dominant 3D perception stack.

Recent point cloud transformers have begun reducing reliance on custom algorithms to better scale self-attention \cite{attention}, though this trend has not directly targeted the broader portability problem.
Because attending to an entire point cloud is computationally intractable, initial point transformers \cite{pointops,ptv2,votr,pointformer,STFormer,sphereformer} form local attention windows through neighborhood grouping (\eg, K-Nearest Neighbors) or volumetric grouping (\eg, radius search, voxel windows).
Neighborhood grouping is memory-inefficient, while volumetric grouping produces windows of unequal point density, leading to irregular parallelization and poor hardware utilization.
Fixed-sequence transformers \cite{flatformer,ptv3,octformer} address these issues with the goal of better model scaling by sorting or unfolding point clouds into equal-length patches, trading precise spatial locality for computational regularity and scaling efficiency.
This allows self-attention to be implemented without external libraries and allows optimizations such as FlashAttention \cite{FA2}.
However, these architectures remain hybrid, retaining sparse 3D convolution \cite{SparseConvNet} and dedicated scatter/gather libraries \cite{PYG} that limit portability to embedded and non-NVIDIA hardware.

We introduce \textbf{PointTransformerX} (PTX), a PyTorch-native point cloud transformer that eliminates all reliance on custom CUDA operators and external libraries, establishing a fully portable transformer backbone for 3D point clouds across diverse hardware targets.
PTX builds on the serialization-based architecture of PointTransformer V3 (PTv3) \cite{ptv3}, but takes the approach further by completely removing sparse algorithms, resulting in a lightweight network with only 20.8\% of PTv3's parameter count.
To maintain competitive accuracy, we make three targeted contributions.
First, we replace sparse convolution with 3D-GS-RoPE and a standard linear projection for patch embedding.
3D-GS-RoPE is a novel Rotary Position Embedding \cite{rope1d} that avoids materializing neighborhood matrices and allows each attention head to independently learn an orthonormal coordinate rotation, enabling cross-axis positional relationships without custom operators.
Second, we show that inference-time attention window scaling is beneficial in conjunction with 3D-GS-RoPE, improving accuracy while reducing the window size during training.
Third, we redesign the feed-forward network (FFN) within the transformer block to further improve efficiency.
Together, these contributions yield a model that achieves $98.7\%$ of PTv3's accuracy on ScanNet \cite{scannet} at just 9.6M parameters and 253\,MB memory (BF16+FP8), effectively matching PTv3 performance without hard dependencies on specialized libraries or CUDA kernels.
PTX is implemented fully in the standard PyTorch ecosystem and runs natively on NVIDIA, AMD, and CPU hardware.

\section{Related Work}

\subsection{Point Cloud Backbones}
Point cloud backbones fall into three main families: point-based, voxel-based, and projection-based methods \cite{revisiting-pc}.
Point-based methods such as PointNet++ \cite{pointnet++}, KPConv \cite{kpconv}, and Edge Convolution \cite{dgcnnh} operate on unstructured point sets and rely on kNN or radius search for local aggregation.
These approaches require $O(N^2)$ pairwise distance computations, which makes them prohibitively expensive at high resolutions.
Voxel-based methods face a similar limitation since naive convolution or attention on dense voxel grids of resolution $R$ scales as $O(R^3)$.
Because most voxels are empty, methods such as SECOND \cite{spconv}, MinkUNet \cite{Minkowski}, and StratifiedTransformer \cite{STFormer} employ sparse algorithms that operate only on occupied voxels to reduce computational and memory costs.
Projection-based methods such as BEV pillar encodings \cite{pointpillars} use standard 2D convolution but incur irreversible geometric information loss.
Furthermore, projection and unprojection as well as pooling and interpolation often rely on custom scatter and gather operators for efficiency \cite{radarpillars}.
Across all families, spatial algorithms heavily depend on custom CUDA operators across a fragmented ecosystem with limited support for modern compute and memory optimizations available in PyTorch.

\subsection{Point Cloud Transformers}

As in other spaces, transformer based backbones see widespread adoption for point cloud perception.
Early point cloud transformers \cite{pointformer,pointops,ptv2,STFormer,sphereformer,votr} used kNN or volumetric grouping for local window attention, producing irregular workloads incompatible with efficient attention implementations and limiting the receptive field.
More recently, fixed-sequence transformers have emerged as a scalable alternative.
These methods impose a one-dimensional ordering on the point cloud and divide the resulting sequence into fixed-length patches.
OctFormer \cite{octformer} adopts the z-order traversal used in octree construction for one-dimensional sequence ordering.
FlatFormer \cite{flatformer} unfolds BEV pillars, while PointTransformer V3 (PTv3) \cite{ptv3} serializes voxels in three-dimensional space using z-order \cite{zcurve} and Hilbert \cite{hilbertcurve} space-filling curves, partitioning the sorted voxels into fixed-size patches processed with window attention.
PTv3 and FlatFormer further apply shuffle-order strategies to improve inter-patch connectivity across layers, whereas OctFormer uses dilated attention.
Since no total ordering can preserve exact three-dimensional neighborhood relationships, this design trades precise spatial locality for computational efficiency.
In PTv3, this increases the receptive field from 16 to 1024 points while achieving $3\times$ higher throughput and $10\times$ lower memory usage compared to its kNN-based predecessor PointTransformer V2 (PTv2) \cite{ptv2}.
OctFormer, FlatFormer, and PTv3 therefore exhibit a consistent reduction in custom operators along the attention pathway.
However, OctFormer and PTv3 still rely on sparse convolution for positional encoding and patch embedding, which results in a hybrid pipeline dependent on external libraries, while FlatFormer remains limited to two-dimensional projected point clouds.
OctFormer is notable in that it provides PyTorch fallbacks to their CUDA kernels, allowing for portability, at the cost of efficiency.

\subsection{Coordinate Position Embedding}
Point cloud transformers typically use relative position embedding (RPE) based on pairwise distances between points \cite{STFormer,pointops,swin3d}.
However while spatially precise, the $O(N^2)$ complexity of RPE is computationally expensive and is a primary factor in preventing attention window scaling.
In contrast, fixed-sequence transformers such as OctFormer and PTv3 rely on hybrid sparse convolution architectures which provide an implicit form of conditional position embedding \cite{cpe}, whereas FlatFormer employs a sinusoidal position embedding defined on the BEV grid.
Rotary Position Embedding (RoPE) \cite{rope1d} has demonstrated strong accuracy gains and computational efficiency in natural language processing, however has not seen adoption for point cloud perception.
It was originally proposed for one-dimensional sequences and later extended to the multidimensional case through Axial-RoPE \cite{axialrope-1,axialrope-2} by partitioning the embedding dimension $d$ into equal subspaces per axis and applying standard one-dimensional RoPE to each subspace.
Concurrent with this work, LitePT \cite{litept} adopts standard Axial-RoPE for point cloud processing.
In contrast, PTX goes one step further by introducing our novel 3D-GS-RoPE, a per-head learned orthonormal coordinate rotation applied before Axial-RoPE that enables cross-axis sensitivity.

\section{Method}

Point cloud networks are typically coupled to custom CUDA operators, restricting deployment to high-end NVIDIA hardware and preventing use of standard PyTorch optimizations on embedded and non-NVIDIA platforms.
These limitations motivate the design of PointTransformerX, whose architectural design goal is to implement the entire network using only native PyTorch operators, thereby removing vendor-specific dependencies while retaining state-of-the-art performance.

\subsection{Preliminary: PointTransformer V3}
We build on PointTransformer V3 (PTv3) \cite{ptv3}, which represents the current state of the art in point cloud processing.
The PTv3 architecture follows a hierarchical U-Net-style \cite{u-net} encoder-decoder with multiple stages, each composed of identical blocks.
Each block consists of a 3D convolutional layer with skip connection, followed by a pre-normalization transformer block.
The transformer block uses window attention and a feed-forward network with expansion ratio $4$ and GeLU \cite{gelu} activation.
Positional embedding is provided implicitly by the convolutional layer.
Within each block, voxels are sorted and partitioned into patches for fixed-sequence attention, alternating between z-Order \cite{zcurve}, translated z-Order, Hilbert-curve \cite{hilbertcurve}, and translated Hilbert-curve to foster cross-patch interaction.
PTv3's sparse convolution accounts for approximately $67\%$ of parameters and requires external libraries for both sparse convolution and scatter/gather operations, forming the primary portability bottleneck in PTv3.

\begin{figure*}[t!]
    \centering
    \includegraphics[width=0.175\linewidth]{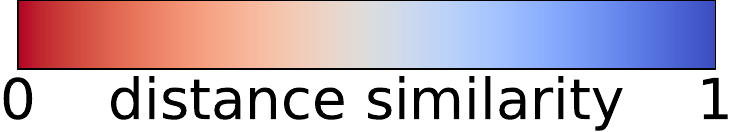} \\
    \begin{subfigure}[b]{0.5\columnwidth}
        \includegraphics[width=\columnwidth]{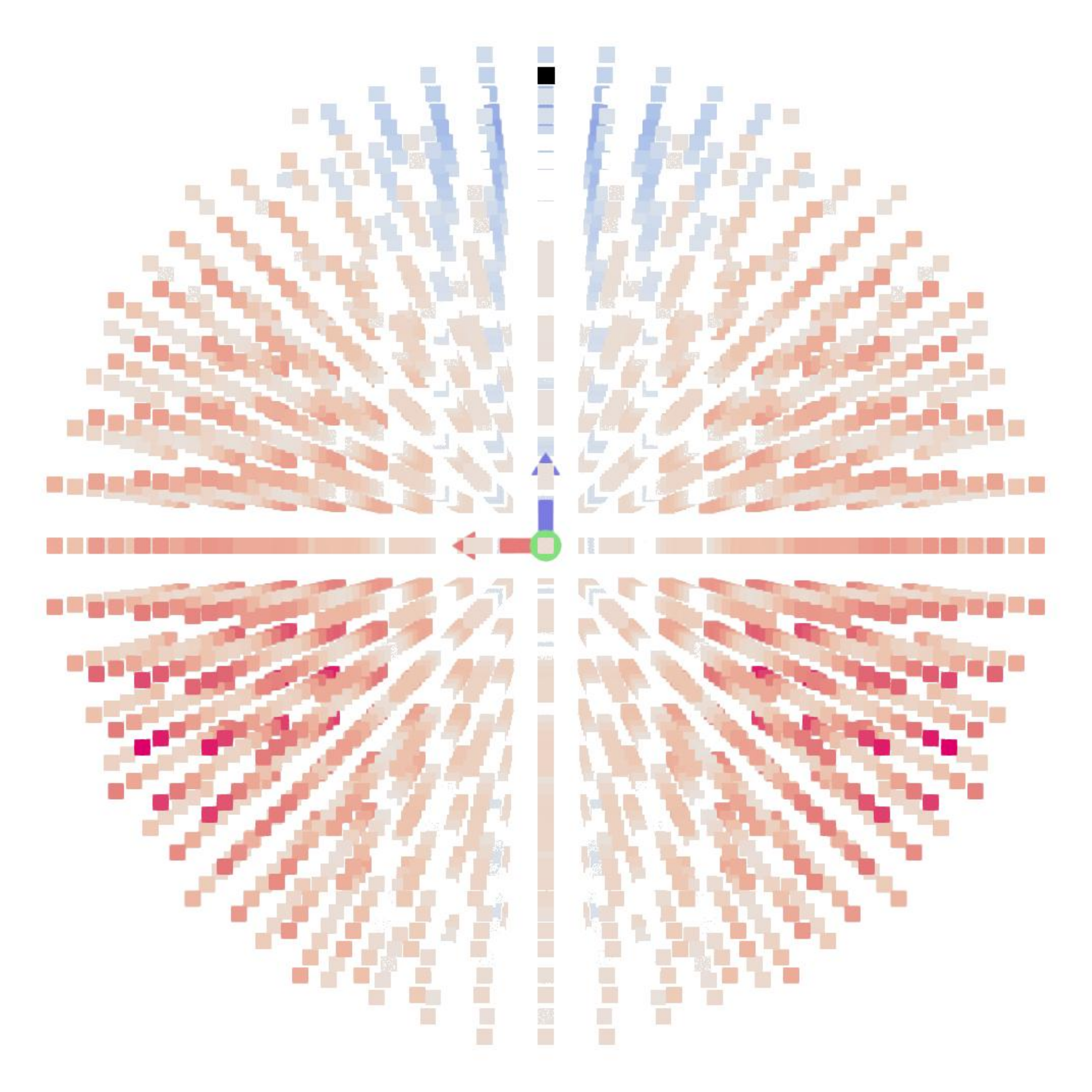}
        \caption*{Axial-RoPE}
    \end{subfigure}
    \hfill
    \begin{subfigure}[b]{0.5\columnwidth}
        \includegraphics[width=\columnwidth]{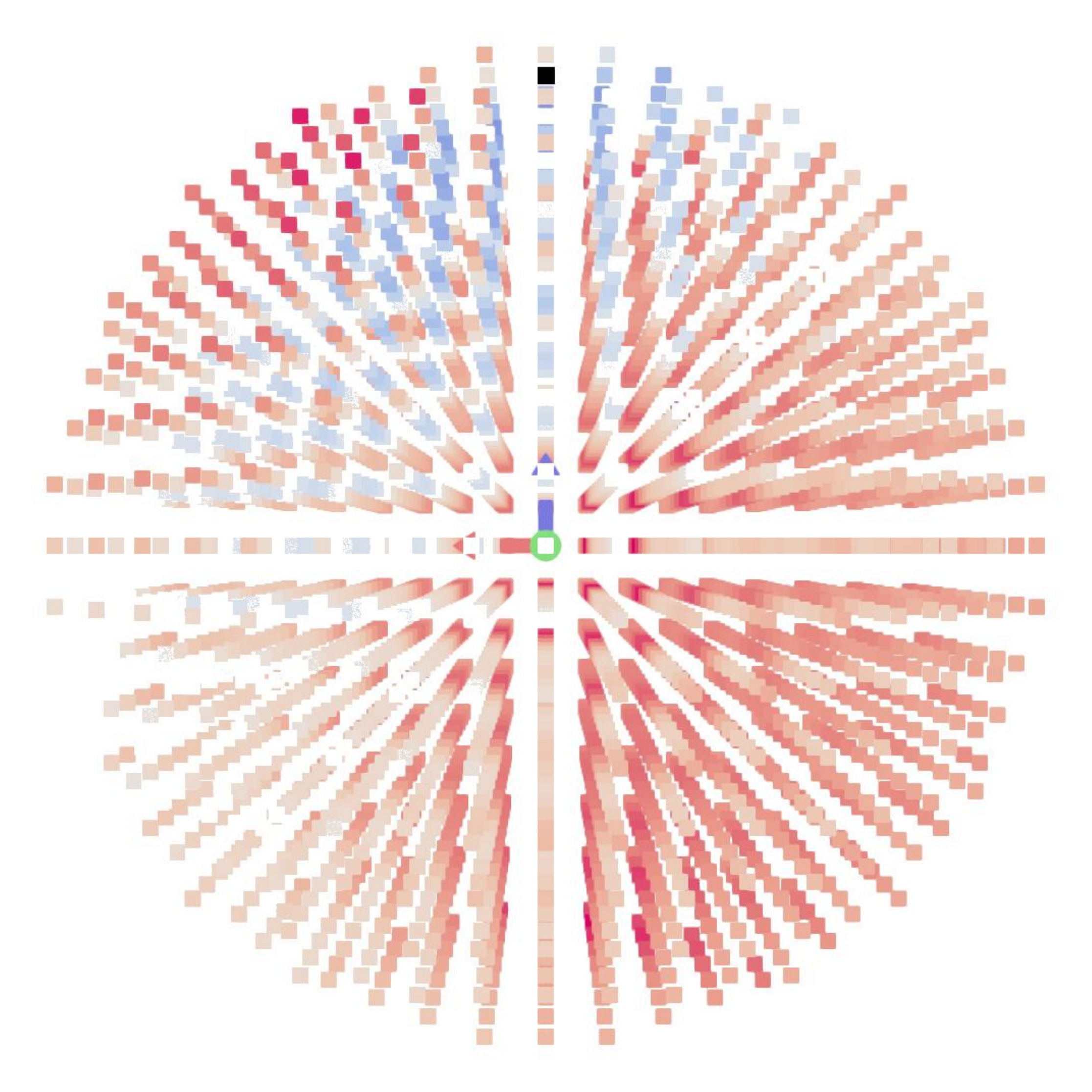}
        \caption*{Mixed-RoPE}
    \end{subfigure}
    \hfill
    \begin{subfigure}[b]{0.5\columnwidth}
        \includegraphics[width=\columnwidth]{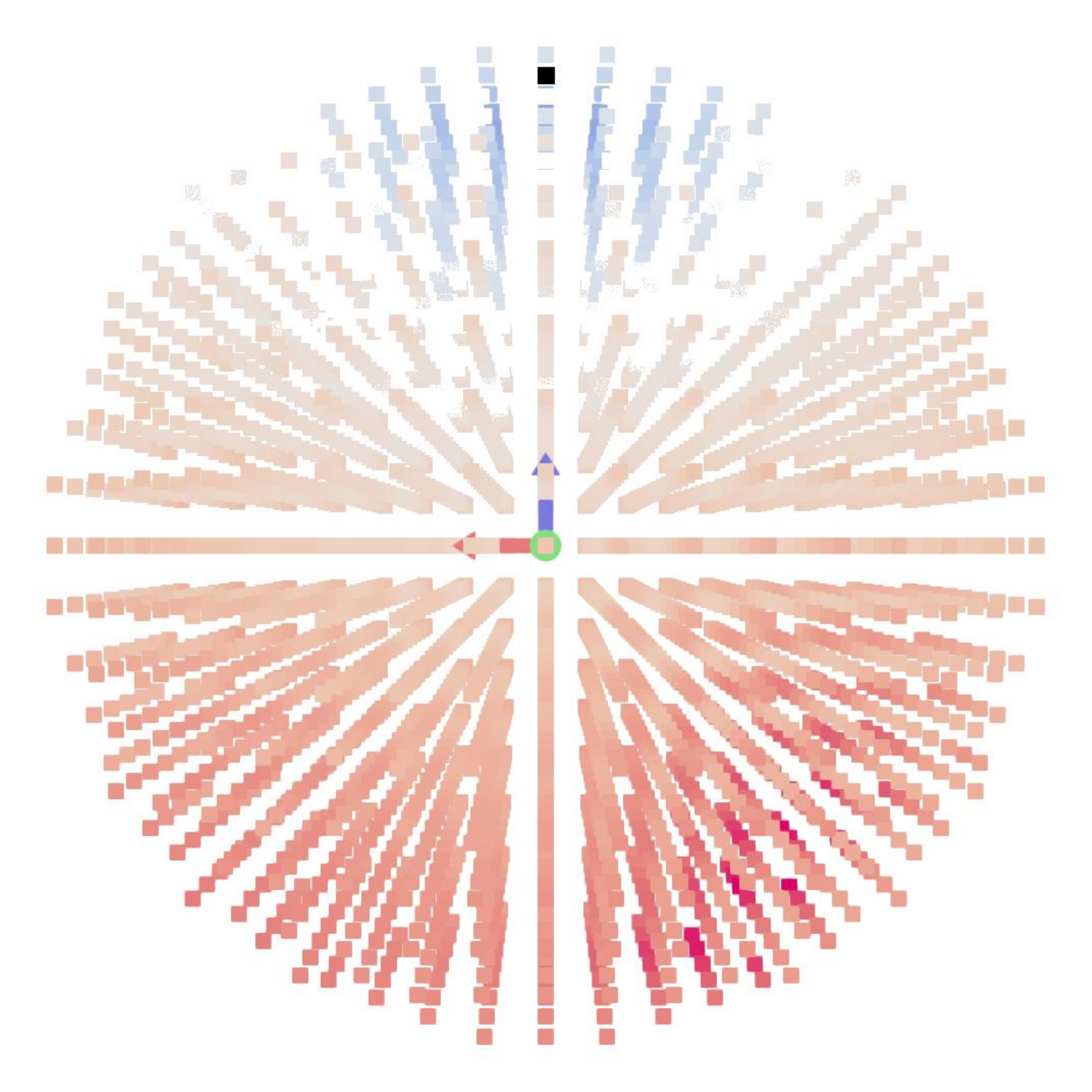}
        \caption*{3D-GS-RoPE}
    \end{subfigure}
    \hfill
    \begin{subfigure}[b]{0.5\columnwidth}
        \includegraphics[width=\columnwidth]{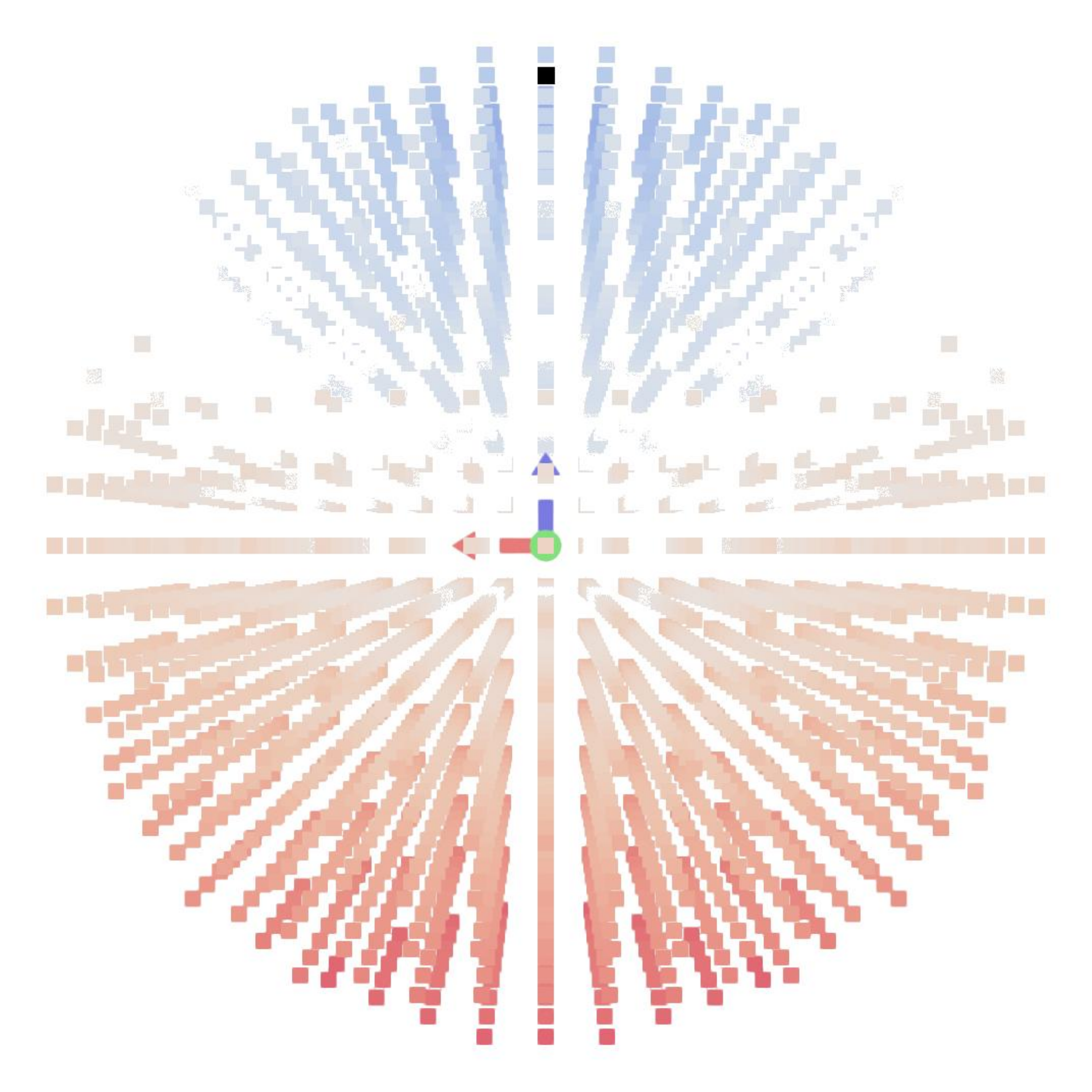}
        \caption*{Euclidean (ground truth)}
    \end{subfigure}
    \vspace*{-0.5\baselineskip}
    \caption{Visualization of spatial relationships modeled by Axial- \cite{axialrope-1,axialrope-2},
        Mixed- \cite{mixedrope}, and our 3D-GS-RoPE for points sampled from a unit sphere
        relative to the query point at $(x=0, y=1, z=0)$.
        The rightmost image shows the Euclidean distance ideal.
        Axial-RoPE lacks cross-axis sensitivity, causing axis-aligned distant points to appear
        more similar to the query than closer diagonal points.
        Mixed-RoPE provides no structural guarantee and encodes spatial relationships
        quasi-randomly at initialization.
        3D-GS-RoPE effectively encodes spatial structure via Gram-Schmidt constraints.
        Mixed- and 3D-GS-RoPE are randomly initialized, while all RoPE variants are shown with $d{=}64$ and $h{=}4$ heads.
        }
    \label{fig:RoPE-comparisons}
\vspace*{-1.25\baselineskip}
\end{figure*}

\subsection{Replacing Sparse Convolution through Rotational Position Embedding}

Removing sparse convolution necessitates explicit positional encoding, as self-attention is permutation invariant and requires geometric context to learn spatial relationships within an unordered point set.
Our objective is to recover PTv3-level accuracy within standard self-attention.

\textbf{RoPE.}
Rotary Position Embedding (RoPE) \cite{rope1d} encodes relative positions so that the inner product between query $q'_n$ and key $k'_m$ tokens depends only on their relative position $m{-}n$, effectively encoding pairwise distance through rotation.
For this, each embedding is split into 2D pairs and rotated proportionally to the token position, which can be expressed as complex multiplication via Euler's formula.
\begin{equation}
  q'_n = q_n e^{i \theta n}, \quad k'_m = k_m e^{i \theta m}
  \label{eq:RoPE:q-k-rotated}
\end{equation}
RoPE utilizes multiple rotation frequencies $\theta_t$ in Eq. \ref{eq:RoPE:q-k-rotated}, one for each of the $d_h/2$ embedding pairs in the head dimension.
\begin{equation}
  \theta_t = \theta_\mathrm{base}^{-t / (d_h/2)},\quad t \in \{0, 1, \ldots, d_h/2 - 1\}
  \label{eq:RoPE:frequencies}
\end{equation}
The full rotation of Eq. \ref{eq:RoPE:q-k-rotated} applied to a token at position $n$ can then be written as
\begin{equation}
  \textbf{R}(n) = e^{i\theta_t n}
  \label{eq:RoPE:rotation-1d}
\end{equation}
\textbf{Axial-RoPE.}
Axial-RoPE \cite{axialrope-1,axialrope-2} extends Eq. \ref{eq:RoPE:rotation-1d} to multidimensional inputs by partitioning the $d_h/2$ embedding pairs into equally sized subsets, one per spatial axis, and applying standard 1D RoPE within each subset.
For three dimensions, the head embedding is divided into three groups of $d_h/6$ pairs each. Each group encodes position along a single coordinate axis using its own set of frequencies $\theta_t$ according to Eq. \ref{eq:RoPE:frequencies} with $t \in \{0, 1, \ldots, d_h/6 - 1\}$:
\begin{equation}
\textbf{R}(x,y,z)  =
\begin{bmatrix}
  e^{i\theta_{t}x}\\[4pt]
  e^{i\theta_{t}y}\\[4pt]
  e^{i\theta_{t}z}
\end{bmatrix}
  \label{eq:RoPE:axial}
\end{equation}
where each row corresponds to one of the three subsets of the head embedding.
This preserves the relative-position property along each individual axis as the attention contribution from one axis depends only on the relative distance along that axis.
However, because each axis is encoded independently, the resulting attention score is separable across axes.
This means that two tokens with the same set of per-axis distances but in different spatial arrangements (\eg, offsets $(a, b, c)$ and $(b, a, c)$) can produce identical attention scores.
In practice, Axial-RoPE therefore lacks cross-axis sensitivity, limiting its ability to capture spatial relationships aligned along the diagonals.

\textbf{Mixed-RoPE.}
To address this limitation, Mixed-RoPE \cite{mixedrope} was proposed for the 2D image domain.
Instead of partitioning the embedding into per-axis subsets, Mixed-RoPE uses the full $d_h/2$ pairs and assigns learned frequencies $\theta_t$ per axis to each pair, allowing each pair to jointly encode all axes.
Generalized to 3D coordinates, the rotation becomes
\begin{equation}
\textbf{R}(x,y,z)  =
  e^{i(\theta^{x}_{t}x + \theta^{y}_{t}y + \theta^{z}_{t}z)}
  \label{eq:RoPE:mixed}
\end{equation}
where $\theta^{x}_{t}, \theta^{y}_{t}, \theta^{z}_{t}$ are learnable frequencies that control the mixing of axes for each embedding pair.
While this enables cross-axis interaction, the unconstrained learning of per-axis frequencies introduces a risk.
If frequencies become imbalanced, an axis can dominate the rotation, distorting the spatial relationships encoded in the position embedding.
Although a network could in principle learn balanced frequencies, there is no structural guarantee that it will.

\textbf{3D Gram-Schmidt RoPE.}
We propose 3D-Gram-Schmidt-RoPE (3D-GS-RoPE), which retains the axial structure of Axial-RoPE but introduces a learned orthonormal coordinate rotation per attention head.
Concurrent work LitePT \cite{litept} differs in that it adopts standard Axial-RoPE, but does not address the cross-axis issue.
Rather than relying on unconstrained frequency learning to achieve cross-axis interaction, 3D-GS-RoPE applies a rigid head-wise rotation to the input coordinates before encoding them with Axial-RoPE.
Because an orthonormal rotation preserves distances and angles, the geometric relationships between points are maintained in the transformed coordinate frame.
Each head learns a different rotation, so the axis-aligned limitations of Axial-RoPE differ across heads, collectively restoring cross-axis sensitivity without introducing additional spatial operators.

Axial-RoPE corresponds to a coordinate transform with the identity matrix, which we generalize from Eq. \ref{eq:RoPE:axial} into a learned orthonormal transform:
\begin{equation}
\begin{aligned}
\textbf{R}(x,y,z)
&=
\begin{bmatrix}
  e^{i\theta_{t}x}\\[4pt]
  e^{i\theta_{t}y}\\[4pt]
  e^{i\theta_{t}z}
\end{bmatrix}
=
\begin{bmatrix}
  e^{i\theta_{t}(1 \cdot x + 0 \cdot y + 0 \cdot z)}\\[4pt]
  e^{i\theta_{t}(0 \cdot x + 1 \cdot y + 0 \cdot z)}\\[4pt]
  e^{i\theta_{t}(0 \cdot x + 0 \cdot y + 1 \cdot z)}
\end{bmatrix}
\\[6pt]
&=
\begin{bmatrix}
  e^{i\theta_{t}(r_{11}x + r_{12}y + r_{13}z)}\\[4pt]
  e^{i\theta_{t}(r_{21}x + r_{22}y + r_{23}z)}\\[4pt]
  e^{i\theta_{t}(r_{31}x + r_{32}y + r_{33}z)}
\end{bmatrix}
\end{aligned}
\label{eq:RoPE:axial-as-transform}
\end{equation}
where $\theta_t$ with $t \in \{0, 1, \ldots, d_h/6 - 1\}$ are the per-group frequencies, and the rows $\textbf{r}_1, \textbf{r}_2, \textbf{r}_3$ of the coordinate transform matrix are

\begin{equation}
\begin{bmatrix}
  \textbf{r}_1 \\
  \textbf{r}_2 \\
  \textbf{r}_3
\end{bmatrix} =
\begin{bmatrix}
  (r_{11}, r_{12}, r_{13}) \\
  (r_{21}, r_{22}, r_{23}) \\
  (r_{31}, r_{32}, r_{33}) \\
\end{bmatrix} =
\begin{bmatrix}
  (1,0,0) \\
  (0,1,0) \\
  (0,0,1) \\
\end{bmatrix} = I_3
  \label{eq:RoPE:identity-case}
\end{equation}

In 3D-GS-RoPE, we allow $\textbf{r}_1, \textbf{r}_2, \textbf{r}_3$ and $\theta_t$ to be learned, constrained to form an orthonormal basis via Gram-Schmidt orthonormalization \cite{gs1,gs2}.
Concretely, $\textbf{r}_1$ is a freely learned vector that is normalized to unit length, $\textbf{r}_2$ is learned and orthonormalized with respect to $\textbf{r}_1$, and $\textbf{r}_3$ is fully determined as the cross product $\textbf{r}_1 \times \textbf{r}_2$.
This adds only six learnable parameters per head (three components for $\textbf{r}_1$ and three for $\textbf{r}_2$, before orthonormalization) plus learnable frequencies $\theta_t$, resulting in negligible overhead relative to sparse convolution.
Since all three vectors have unit norm by construction, projecting a point coordinate $\textbf{p}_i = (x, y, z)$ onto the rotated axes simplifies to
\begin{equation}
x' = \mathbf{r}_1^\top \mathbf{p}_i,\quad
y' = \mathbf{r}_2^\top \mathbf{p}_i,\quad
z' = \mathbf{r}_3^\top \mathbf{p}_i
\label{eq:RoPE:projection}
\end{equation}
Applied to Eq. \ref{eq:RoPE:identity-case} the final rotation applied to a head embedding becomes
\begin{equation}
\textbf{R}(x,y,z) =
\begin{bmatrix}
  e^{i\theta_{t}\,\mathbf{r}_1^\top \mathbf{p}_i}\\[4pt]
  e^{i\theta_{t}\,\mathbf{r}_2^\top \mathbf{p}_i}\\[4pt]
  e^{i\theta_{t}\,\mathbf{r}_3^\top \mathbf{p}_i}
\end{bmatrix}
\label{eq:RoPE:gs-rope}
\end{equation}
Each attention head learns its own orthonormal basis $\{\textbf{r}_1, \textbf{r}_2, \textbf{r}_3\}$, resulting in a different rotated coordinate frame per head.
Between attention heads, we cyclically permute the order in which the Gram-Schmidt constraints are applied, using
$\{\mathbf{r}_1 \rightarrow \mathbf{r}_2 \rightarrow \mathbf{r}_3\}$,
$\{\mathbf{r}_2 \rightarrow \mathbf{r}_3 \rightarrow \mathbf{r}_1\}$, and
$\{\mathbf{r}_3 \rightarrow \mathbf{r}_1 \rightarrow \mathbf{r}_2\}$.
This ensures that no single axis is consistently treated as the primary unconstrained direction, preventing bias toward any particular basis vector.

\cref{fig:RoPE-comparisons} shows the distance similarity of a query point to its spatial neighbors on the unit sphere, as encoded by Axial-, Mixed-, and 3D-GS-RoPE.
The lack of cross-axis sensitivity in Axial-RoPE is clearly evident, with points diagonal to the query point showing less similarity in comparison to those aligned with the axes.
The lack of structural guarantee to maintain axis balancing in Mixed-RoPE is also clearly evident.
Here, both Mixed- and 3D-GS-RoPE are initialized randomly.
The Gram-Schmidt constraint more accurately models spatial relationships by reducing the cross-axis effect of Axial-RoPE, approaching Euclidean structure while remaining fully PyTorch-native.

\textbf{Linear Embedding}
We replace the remaining sparse convolutional patch embedding with a standard linear projection layer and remove all custom scatter and gather operations.
This enables a fully PyTorch-native implementation without sacrificing downstream accuracy.

\subsection{Window-Size Inference Scaling}
\label{sec:window-training}

RoPE's position-independent multi-frequency structure, where low frequencies capture long-range relationships and high frequencies capture local context, allows generalization to coordinate ranges unseen during training.
This property is well established in natural language processing \cite{yarn,ntk-1,ntk-2}, where RoPE-based LLMs trained on short sequences can be extended to much longer contexts after training, but has not been studied in the context of point cloud transformers.
We observe this effect through 3D-GS-RoPE and propose scaling the window size at inference beyond the training window size.
Our experiments \cref{sec:experiments} identify a universal scaling factor $s$ for PTX that improves accuracy with negligible training cost, further closing the gap to PTv3 while preserving the sparse-free design.

\subsection{Efficient Feed-Forward Network}
\label{sec:ffn}

PTv3 employs a feed-forward network (FFN) with two linear projections and a non-linear activation in between:
\begin{equation}
\mathrm{FFN}(\mathbf{x}) = \mathbf{x} + W_2 \,\sigma\!\left(W_1 \mathbf{x} + \mathbf{b}_1\right) + \mathbf{b}_2
\label{eq:ffn}
\end{equation}
where $W_1 \in \mathbb{R}^{rd \times d}$ expands the feature dimension by an expansion ratio $r$, and $W_2 \in \mathbb{R}^{d \times rd}$ projects it back to the original dimension. GeLU serves as the activation function, and the expansion ratio is set to $r{=}4$. This configuration originates from the original Transformer and has remained largely unchanged in the point cloud domain.
In contrast, modern large language models have widely adopted SwiGLU \cite{swiglu}, while recent research proposes $\text{ReLU}^2$ \cite{relu2} as a more efficient alternative.

However, point cloud datasets are orders of magnitude smaller than language or image datasets, making the large FFN ($r{=}4$) prone to overfitting.
Since the FFN operates independently per token without contributing spatial interaction, it is a natural target for efficiency optimization, allowing parameter reduction without degrading geometric modeling capacity.
We revisit the FFN design across activation function and expansion ratio. Specifically, we evaluate three activation functions $\sigma$ (GeLU, SwiGLU, and $\text{ReLU}^2$) across expansion ratios $r \in {0.5, 1, 2, 4}$, resulting in 30 configurations. We also experiment with a DenseNet \cite{densenet} style skip connection, increasing the hidden dimension to $rd + d$ and enabling the second linear layer to access richer features, while maintaining the overall sparse-free and hardware-agnostic architecture.

\section{Experiments}\label{sec:experiments}
We follow the architecture and training settings of PointTransformer V3 (PTv3) \cite{ptv3}.
The model follows PTv3's hierarchical U-Net encoder-decoder with 5 encoder stages (depths $2,2,2,6,2$; channels $32,64,128,256,512$) and $4$ decoder stages (depth $2$, channels $256,128,64,64$).
We employ $(2, 4, 8, 16, 32)$ heads in the encoder and $(16, 8, 4, 4)$ in the decoder.
All layers use a window size of 1024 unless otherwise stated.
The model is trained using the AdamW \cite{adamw} optimizer with a learning rate of 0.006 and weight decay of 0.05, scheduled with a period followed by cosine decay \cite{warmupcosine}.
We combine CrossEntropyLoss and LovaszLoss \cite{lovaszloss} as our training objectives for semantic segmentation with equal weighting.
Unlike PTv3, which requires mixed FP16/BF16 precision due to sparse convolution library constraints, PTX uses uniform BF16 throughout.
Points are voxelized at $2$ cm.
Training is performed with a batch size of $12$ on a single NVIDIA H100 on the ScanNet \cite{scannet} dataset with $20$ semantic classes.
We implement our training pipeline in the Pointcept \cite{pointcept} framework.
Final results are reported with test-time augmentation.

\subsection{3D-GS-RoPE Position Embedding}

As the first step toward a PyTorch-native network, we remove all sparse-convolution layers from PTv3, leaving a pure transformer architecture.
We ablate the choice of position embedding, comparing our proposed 3D-GS-RoPE to various RoPE variants.

Our results, shown in \cref{Table:RoPE}, highlight the need for an effective position embedding, as removing sparse convolution results in an accuracy drop of $-12.1\%$.
Standard 1D RoPE can be applied to the fixed-sequence order, but does not preserve spatial information.
Expanding this into Axial-RoPE recovers $96.6\%$ of the original accuracy.
In contrast to 3D-GS-RoPE and Axial-RoPE, Mixed-RoPE performs worse, which can be attributed to a learned imbalance as a result of mixing coordinate axes through $\theta_x, \theta_y$ and $\theta_z$ and evident in the model weights.
We also experiment with partial 3D-GS-RoPE, where rotation is not applied across the entire embedding dimension $d$.
Applying partial-RoPE \cite{partialrope} only to the last encoder stage with $C=256$ improves accuracy slightly, recovering $97.2\%$ of baseline accuracy.

\subsection{Linear Embedding}

\begin{table}[t]
\caption{
Evaluation of RoPE-based position embeddings and their effectiveness in replacing the inductive bias of sparse convolution.
}
\centering
\resizebox{0.85\columnwidth}{!}{
\begin{tabularx}{\columnwidth}{Xcc}
\hline
Position Embedding                                                     & mIoU            & Params (M) \\ \hline
PTv3 baseline* & \textbf{0.7560} & {\ul 46.1}           \\
-- Convolution                                                          & 0.6641          &   \textbf{ 14.2}              \\
+ RoPE-Variants:  & &  \textbf{ 14.2} \\
\qquad 1D RoPE                                                                & 0.6791          &                \\
\qquad Axial RoPE                                                             & 0.7308          &          \\
\qquad Mixed RoPE                                                             & 0.6876          &                \\
\qquad  3D-GS-RoPE                                                             & 0.7331          &                \\
\qquad \begin{tabular}[c]{@{}l@{}}3D-GS-RoPE\\ + partial RoPE\end{tabular}    & {\ul 0.7351}    &                \\ \hline
\multicolumn{3}{l}{* Re-implemented}                                                                     
\end{tabularx}}
\label{Table:RoPE}
\end{table}
\begin{table}[t]
\centering
\caption{
By trading performance and replacing sparse convolution with a linear layer for input embedding, PointTransformerX can be fully implemented efficiently with PyTorch native operators.
}
\resizebox{\columnwidth}{!}{
\begin{tabular}{lccc}
\hline
Embedding Layer & mIoU   & Params (M) & PyTorch-native \\ \hline
Convolution & \textbf{0.7351} & 14.2  &     \ding{55}        \\
Linear & {\ul 0.7232} & 14.2  &    \ding{51}     \\
mini-PointNet & 0.7225 & 14.2   &     \ding{51}       \\ \hline
\end{tabular}
}
\label{Table:Embedding}
  \vspace*{-\baselineskip}
\end{table}

Continuing with our goal of a PyTorch-native network, we replace the sparse-convolutional embedding (which uses a stride of one) with a simple linear layer.
We also experiment with a PointNet-style \cite{pointnet} MLP made up of three linear layers.
Other common embedding operators require custom CUDA operators and are not studied as they defeat the goal of PTX.
The results of this change are depicted in \cref{Table:Embedding}.
At a slight cost of accuracy, this change in conjunction with replacing the custom scatter and gather operations with PyTorch operators results in a PyTorch-native network and unlocks the associated efficiency improvements.

\subsection{Window Size Scaling}\label{abl:wininf}
Having removed all sparse convolutions and their associated inductive biases, we investigate window size scaling during inference.
We observe that training with a reduced window size of $512$ and scaling to $2048$ during inference yields superior accuracy compared to curriculum learning approaches.
\cref{tab:reg-window-training-inference-scaling} shows different training configurations and we consistently find an optimal inference scaling factor of $s=4$.
Notably, we find that the first encoder stage does not benefit from scaling and therefore do not scale the attention in that layer.
We use this setting for the remaining experiments.

\begin{table}[t]
  \centering
  \caption{
  Inference scaling the attention window size after training with a fixed window size consistently improves results.
  Colors reflect the improvement relative to the baseline setting in each column, which trains and evaluates with the same window size.
  }
  \includegraphics[width=1.0\columnwidth]{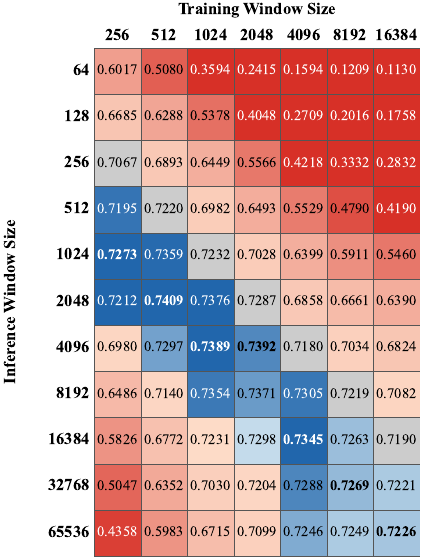}
  \vspace*{-2\baselineskip}
  \label{tab:reg-window-training-inference-scaling}

\end{table}

We hypothesize this improvement arises because the network specializes to its training window size, acquiring beneficial local inductive biases.
We tested curriculum learning that linearly increases the window size during training to prevent overspecialization to the window size.
However, training with a curriculum up to a window size of $2048$ reduces this benefit, achieving only $0.729$ mIoU compared to $0.741$ mIoU with window size inference scaling.
Inference scaling therefore not only compensates for the removal of sparse convolution, but slightly surpasses the fixed-window baseline.
We also tested YaRN~\cite{yarn} scaling but observed no improvement.
This is likely attributable to two factors.
First, patches of serialized point clouds exhibit varying volumes and coordinate
distributions, exposing RoPE frequencies to a broad range of coordinates during training.
Second, YaRN frequency compression assumes certain dimensions are out-of-distribution,
whereas here they are already well-conditioned by the continuous spatial coordinates.

\begin{table*}[t!]
  \centering
  \caption{
  Impact of our contributions on PointTransformerX's accuracy, parameter count, average runtime and average peak allocated memory for ScanNet \cite{scannet} in comparison to the baseline PointTransformer V3 \cite{ptv3} on an NVIDIA H100.
  }
  \vspace*{-\baselineskip}
  \includegraphics[width=2.0\columnwidth]{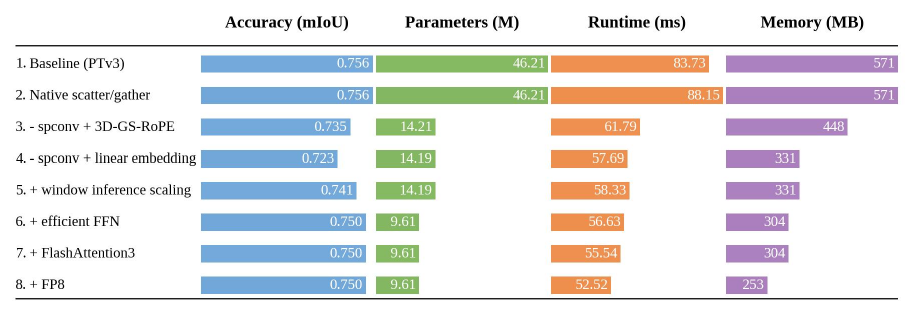}
  \vspace*{-\baselineskip}
  \label{tab:component-analysis}
\end{table*}

\subsection{Efficient FFN}

Our final contribution is the redesign of the standard feed-forward network (FFN) within each transformer blocks.
We ablate the choice of activation function (GeLU, SwiGLU and $\mathrm{ReLU}^2$), the embedding dimension ratio $r$, as well as a DenseNet style skip connection to artificially increase the hidden dimension.
Results are shown in \cref{Table:EfficientFFN}.

$\mathrm{ReLU}^2$ consistently outperforms GeLU in all but the standard $r=4$ setting, while being simpler than SwiGLU.
The use of a DenseNet-style skip connection (DN) adds additional computational and memory overhead.
As the adoption of DN is not conclusive across variants, we utilize a $r=2$ and $\mathrm{ReLU}^2$ configuration for our final network, reducing parameters while preserving PTv3-level accuracy.

\begin{table}[t]
\centering
\caption{
Ablation of the FFN design, varying the expansion ratio $r \in \{1, 2, 4\}$, activation function (GeLU, SwiGLU \cite{swiglu}, $\mathrm{ReLU}^2$ \cite{relu2}), and the use of DenseNet (DN) \cite{densenet} skip connections. 
$\mathrm{ReLU}^2$ improves performance consistently, while 
DN skip connections yield mixed results at added complexity. 
}
\resizebox{0.75\columnwidth}{!}{%
\begin{tabular}{ccccc}
\hline
\multirow{2}{*}{\begin{tabular}[c]{@{}c@{}}Ratio\\ $r$\end{tabular}} &
\multirow{2}{*}{\begin{tabular}[c]{@{}c@{}}Activation\\ Function\end{tabular}} &
\multicolumn{2}{c}{mIoU} \\
\cline{3-4}
 & & w/o DN Skip & w/ DN Skip \\
\hline
\multirow{3}{*}{0.5} & GeLU              & 0.7362 & 0.7299 \\
                     & SwiGLU            & 0.7409 & 0.7329 \\
                     & $\mathrm{ReLU}^2$ & 0.7468 & 0.7399 \\ \hline
\multirow{3}{*}{1}   & GeLU              & 0.7392 & 0.7305 \\
                     & SwiGLU            & 0.7416 & 0.7456 \\
                     & $\mathrm{ReLU}^2$ & 0.7469 & 0.7334 \\ \hline
\multirow{3}{*}{2}   & GeLU              & 0.7458 & 0.7359 \\
                     & SwiGLU            & 0.7388 & 0.7485 \\
                     & $\mathrm{ReLU}^2$ & {\ul 0.7495} & \textbf{0.7503} \\ \hline
\multirow{3}{*}{3}   & GeLU              & 0.7381 & 0.7294 \\
                     & SwiGLU            & 0.7407 & 0.7343 \\
                     & $\mathrm{ReLU}^2$ & 0.7405 & 0.7416 \\ \hline
\multirow{3}{*}{4}   & GeLU              & 0.7409 & 0.7311 \\
                     & SwiGLU            & 0.7440 & 0.7353 \\
                     & $\mathrm{ReLU}^2$ & 0.7349 & 0.7469 \\ \hline
\end{tabular}%
}
\label{Table:EfficientFFN}
  \vspace*{-\baselineskip}
\end{table}

\subsection{Results}

\begin{table*}[t!]
\centering
\resizebox{2\columnwidth}{!}{
\renewcommand{\arraystretch}{1.15}
\footnotesize
\setlength{\tabcolsep}{5pt}
\begin{threeparttable}
\caption{
Comparison of point cloud segmentation backbones across hardware platforms. Peak memory and runtimes are averaged over the ScanNet evaluation split \cite{scannet}. CPU refers to Intel Core i5-10400F.
mIoU is reported with test-time augmentation.
PTX uses BF16 on all hardware, except FP8 for applicable linear layers on the NVIDIA H100. For fair comparison, all applicable networks are upgraded to FlashAttention-3 \cite{FA3} on the H100.
}
\label{table:SOTA}
\begin{tabular}{@{}llcrrrrrrrrc@{}}
\toprule
 & & & & & \multicolumn{2}{c}{NVIDIA H100} & \multicolumn{2}{c}{AMD MI300X} & \multicolumn{2}{c}{ Jetson Orin AGX} & {CPU} \\
\cmidrule(lr){6-7} \cmidrule(lr){8-9} \cmidrule(lr){10-11} \cmidrule(l){12-12}
Backbone & Architecture & \makecell{PyTorch\\Native} & \makecell{Params\\(M)} & mIoU & \makecell{Mem.\\(GB)} & \makecell{Rt.\\(ms)} & \makecell{Mem.\\(GB)} & \makecell{Rt.\\(ms)} & \makecell{Mem.\\(GB)} & \makecell{Rt.\\(ms)} & \makecell{Rt.\\(s)} \\
\midrule
PTX (ours)              & Transformer & \cmark & {\ul 9.6}  & {\ul 0.765}   & \textbf{0.253} & {\ul 52.52}   & \textbf{0.390} & \textbf{53.98}   & \textbf{0.239} & 259.54  & 3.40 \\
PTv1 \cite{pointformer} \tnote{3}           & Transformer & \xmark & \textbf{7.8}  & 0.706 & 0.749 & 1036.21 & \nx   & \nx     & 0.722 & 2288.57 & \nx  \\
PTv2 \cite{ptv2} \tnote{3}            & Transformer & \xmark & 11.3 & 0.754 & 1.652 & 173.16  & \nx   & \nx     & 1.625 & 948.41  & \nx  \\
ST \cite{STFormer}                      & Transformer & \xmark & 18.8 & 0.743 & \na   & \na     & \nx   & \nx     & \na   & \na     & \nx  \\
Swin3D-S \cite{swin3d} \tnote{3,4}       & Transformer & \xmark & 28.2 & 0.764 & 0.525   & 81.00     & \nx   & \nx     & 0.803   & 2007.28     & \nx  \\
\addlinespace
LitePT-S \cite{litept} \tnote{2}      & Hybrid      & \xmark & 12.7 & {\ul 0.765} & {\ul 0.314} & 56.81   & \nx   & \nx     & {\ul 0.283} & {\ul 196.34}  & {\ul 1.75} \\
Octformer \cite{octformer} \tnote{1}   & Hybrid      & \cmark & 44.0 & 0.757 & 2.073 & 58.23   & {\ul 1.585} & {\ul 72.33}   & 2.038 & 426.17  & 6.68 \\
PTv3 \cite{ptv3} \tnote{2}            & Hybrid      & \xmark & 46.2 & \textbf{0.775} & 0.562 & 83.73   & \nx   & \nx     & 0.541 & 285.55  & 3.48 \\
\addlinespace
MinkUNet-34C \cite{Minkowski} \tnote{3}    & Convolution & \xmark & 37.9 & 0.722 & 0.525 & 81.00   & \nx   & \nx     & 0.503 & 465.84  & 5.68 \\
SpUNet \cite{spconv} \tnote{2}                  & Convolution & \xmark & 39.2 & 0.693 & 0.487 & \textbf{43.02}   & \nx   & \nx     & 0.464 & \textbf{138.58}  & 1.80 \\
OACNN \cite{oacnn} \tnote{2}           & Convolution & \xmark & 51.5 & 0.761 & 0.708 & 132.89  & \nx   & \nx     & 0.566 & 489.27  & \textbf{1.65} \\
\bottomrule
\end{tabular}
\begin{tabular}{p{0.29\linewidth} p{0.27\linewidth} p{0.26\linewidth} p{0.14\linewidth}}
\textsuperscript{1} CUDA kernels with PyTorch fallback &
\textsuperscript{2} Implementation adapted for CPU &
\textsuperscript{3} Manual update of CUDA kernels &
\textsuperscript{4} Extra data
\end{tabular}
\end{threeparttable}
}
\vspace*{-1\baselineskip}
\end{table*}

\textbf{Component Contributions and Efficiency.}
We analyze the efficiency gains of PTX relative to the PTv3 baseline, with component impact summarized in \cref{tab:component-analysis}.

Initially, replacing custom CUDA operators for scatter/gather operations in the network's pooling and unpooling layers with native PyTorch operators preserves accuracy but detrimentally increases runtime.
The following optimizations recover this regression.
Removing sparse convolution is the most impactful and substantially reduces parameter count, memory use and runtime.
In contrast 3D-GS-RoPE is lightweight, as the decomposition of rotations means that distance matrices are not materialized.

The embedding layer replacement is particularly impactful for memory.
As the point cloud resolution is highest in the first encoder stage, sparse convolutional embedding is especially memory-intensive there, making it the dominant stage for peak memory allocation.
Removing sparse convolution further allows uniform BF16 precision throughout the network, eliminating the frequent FP16/BF16 type conversions required by PTv3 and reducing memory operations.

As discussed in \cref{abl:wininf}, scaling the inference window from $512$ to $2048$ but keeping the first encoder stage window size fixed has the benefit that peak memory allocation remains unchanged, as voxel resolution and memory requirements are highest at this stage. Runtime, however, slightly increases.

Our optimized FFN provides modest improvements in runtime and accuracy, and upgrading from FlashAttention-2 \cite{FA2} to -3 \cite{FA3} offers similar gains.
Further improvements are achieved by partially downcasting linear layers with an embedding dimension divisible by $128$ to Float8.
However, while the improved FFN, FA3 and Float8 provide marginal improvements, profiling reveals that PTX has become highly efficient in GPU compute utilization, making other limitations dominant.
Pooling and unpooling in point cloud networks remain inefficient due to the unordered nature of the data, which requires identifying unique indices for scatter and gather operations.
Both the serialization of point clouds and pooling introduce sequential sorting operations throughout the network, limiting parallelization and causing CPU/GPU synchronization.
This results in steps with high CPU utilization, blocking further operations and capping efficiency gains.

\textbf{Comparison to State of the Art.}
We compare PTX against Transformer-based, convolutional, and hybrid architectures across
accuracy, average peak memory utilization, and average inference runtime, with results reported in \cref{table:SOTA}.
PTX is the only architecture that relies entirely on native PyTorch operators, requiring
no third-party CUDA libraries or custom CUDA kernels, yet it ranks competitively across
most metrics while matching the accuracy of PTv3.
Notably, PTX operates with the largest attention window among all compared methods at
$2048$ tokens, versus $1024$ for PTv3 and LitePT, while consuming only $253$ MB of average peak
memory at BF16/FP8 precision on an H100.
In terms of GPU-accelerated inference, PTX at $52.52$ ms is only surpassed by SpUNet with $43.02$ ms, which lags behind in accuracy.
However, this advantage is directly tied to SpUNet’s use of the \texttt{spconv} \cite{spconv} library, as the \texttt{MinkowskiEngine} \cite{Minkowski} sparse convolution used by MinkUNet is significantly slower. 
We also note that \texttt{spconv}, \texttt{MinkowskiEngine}, and most third-party CUDA kernels used by other methods are no longer actively maintained and required manual updates to run under CUDA $12.6$ on the Hopper architecture.
We also compare runtime and memory on an NVIDIA Jetson Orin AGX using BFloat16 precision.
PTX has the lowest memory consumption at a competitive runtime, showing its potential for embedded hardware.

\textbf{Performance on CPU and AMD Hardware.}
As a PyTorch-native architecture, PTX transfers seamlessly to non-NVIDIA hardware without
modification.
On an AMD Instinct MI300X, PTX achieves an average runtime of $53.98$ ms with a peak
memory allocation of $390$ MB with BF16 precision.
For a fair comparison, we adapted other models to enable CPU inference.
For example, we removed all biases in convolutional layers using the \texttt{spconv} framework, as it does not support bias addition on CPU, and replaced CUDA-specific operations with PyTorch operators for CPU execution.
On an Intel i5 CPU, PTX achieves an average runtime of $3.4$~s under BF16 precision.
Among all compared methods, OctFormer is the only architecture with comparable
portability, offering a PyTorch-native fallback for its CUDA depthwise octree convolution.
However, with this fallback OctFormer incurs the highest memory usage of all evaluated models and has the worst runtime on both CPU and AMD hardware.
These results position PTX in a class of its own as a portable, hardware-agnostic
architecture for 3D perception.
As a further testament to this, PTX is exportable to ONNX \cite{onnx}, verified under opset $18$.
\section{Conclusion}
We introduced PointTransformerX (PTX), a fully portable PyTorch-native transformer backbone for 3D point clouds, eliminating all custom CUDA operators, removing the vendor lock and portability barriers that pervade existing point cloud networks.
PTX addresses these limitations with 3D-GS-RoPE for position encoding, window-size inference scaling, and an efficient feed-forward network design, collectively achieving PTv3-level accuracy without sparse convolution while substantially improving efficiency.
By removing framework and hardware dependencies, PTX establishes a clean transformer-only baseline that generalizes seamlessly across GPU vendors and CPUs.

Experiments revealed remaining areas for future improvement.
Specifically, indexing for pooling and serialization operations introduce sequential sorting, which incurs CPU/GPU synchronizations and introduces operations that are difficult to parallelize.
Token pruning offers an alternative direction, as a large fraction of tokens usually do not contribute to accuracy.
A second approach could be inheriting the serialized order of the point cloud and using strided 1D convolution over spatial pooling for hierarchical aggregation.
Finally, PTX's lightweight transformer-only architecture opens up new possibilities for studying data and architectural scaling.

\section*{Acknowledgements}
This work was partially funded by the German Federal Ministry of Research, Technology and Space (BMFTR) under grant Safe3D (12FH568KX2) and by Pepperl+Fuchs SE.

{
    \small
    \bibliographystyle{ieeenat_fullname}
    \bibliography{main}
}

\end{document}